%% file: acl_latex.tex
\pdfoutput=1

\documentclass[11pt]{article}

\usepackage[preprint]{acl}

\usepackage{times}
\usepackage{latexsym}
\usepackage{amsmath}
\usepackage{amsfonts}
\usepackage{booktabs}

\usepackage[T1]{fontenc}

\usepackage[utf8]{inputenc}

\usepackage{microtype}

\usepackage{inconsolata}

\usepackage{graphicx}
\usepackage{tcolorbox}

\usepackage{listings}
\title{Measuring Intrinsic Dimension of Token Embeddings}

\author{%
  Takuya Kataiwa$^{1,2}$ \quad Cho Hakaze$^{2}$ \quad Tetsushi Ohki$^{1,3}$\\
  $^{1}$Shizuoka University\\ $^{2}$Japan Advanced Institute of Science and Technology \\
  $^{3}$RIKEN AIP \\
  \texttt{kataiwa.takuya.23@shizuoka.ac.jp}, \texttt{yfzhao@jaist.ac.jp},\\ \texttt{ohki@sec.inf.shizuoka.ac.jp}
}

\begin{document}
\maketitle
\begin{abstract}
\input{section/abstract}
\end{abstract}

\section{Introduction}

\input{section/sec1_introdcution}

\section{Related Works}

\input{section/sec2_related}

\section{Methodology and Experiments}

\input{section/sec3_method}

\section{Discussions}

\input{section/sec4_discussion}

\section{Conclusion}
\input{section/sec5_conclusion}

\newpage

\section*{Limitation}
\input{section/limitation}



\bibliography{custom}

\newpage

\appendix
\section{Tokenizer Analysis}
\label{sec:appendix}

\input{section/Appendix}

\end{document}

%% file: section/abstract.tex
In this study, we measure the Intrinsic Dimension (ID) of token embedding to estimate the intrstic dimensions of the manifolds spanned by the representations, so as to evaluate their redundancy quantitatively compared to their extrinsic dimensionality. In detail, (1) we estimate the ID of token embeddings in small-scale language models and also modern large language models, finding that the embedding spaces often reside on lower-dimensional manifolds compared to their extrinsic dimensionality; (2) we measure the ID across various model sizes and observe an increase in redundancy rates as the model scale grows; (3) we measure the dynamics of IDs during the training process, and find a rapid ID drop in the early stages of training. Moreover, (4) when LoRA is applied to the embedding layers, we observe a sudden drop in perplexity around the estimated IDs, suggesting that the ID can serve as a useful guideline for LoRA application.



%% file: section/sec1_introdcution.tex
Recent \textbf{L}arge \textbf{L}anguage \textbf{M}odels (LLMs) utilize token embedding layers with hundreds or even thousands of \emph{\textbf{e}xtrinsic \textbf{d}imensions} (ED), while it remains unclear how many of these dimensions are actually necessary for effective representation. If the token embedding utilizes only a lower-dimensional manifold, large portions of the parameter space may be redundant, increasing training and inference costs unbeneficially. Also, prior work suggests that \textbf{sentence} embeddings can lie on remarkably low-dimensional manifolds~\cite{ueda2024measuring}, while the sentence embeddings are model outputs, or \textit{activations} that can not be explicitly reduced for a more efficient model. 

So, in this paper, we focus on the \textbf{token} embedding, which is the model parameters on the first layer of a typical language model, instead of activations. In detail, we examine the \textbf{Intrinsic Dimension (ID)} of embedding spaces in both small-scale (e.g., Word2Vec, GloVe) and large-scale (e.g., Pythia) word embedding models, addressing two central research questions:
\begin{tcolorbox}
\vspace{-0.2\baselineskip}
\begin{description}
    \item[RQ1] How large is the gap between ED and ID, and what factors influence it?
    \vspace{-0.45\baselineskip}
    \item[RQ2] How does the ID in an LLM’s embedding layer evolve and stabilize among model scale and training dynamics?
\end{description}
\vspace{-1\baselineskip}
\end{tcolorbox}

To answer these questions, first, we measure the discrepancy between ED and ID in popular word embedding models (Section~\ref{subsec:exp1}). Next, using Pythia suite~\cite{biderman2023pythia}, we investigate how the dimension redundancy varies against model scales, and how the IDs update among the training dynamics (Sections~\ref{subsec:exp2} and~\ref{subsec:exp3}). Finally, we show that the estimated ID can guide the selection of the inner dimension in \textbf{lo}w-\textbf{r}ank \textbf{a}daptation (LoRA)~\cite{hulora} on the embedding layer, striking a better balance between compactness and performance (Section~\ref{subsec: LoRA}).

\paragraph{Contributions.}
(1) We present a consistent empirical analysis of ID for both small- and large-scale embedding models, demonstrating that embedding spaces remain surprisingly low-dimensional.  
(2) We reveal that the ID is stabilized in the early training even as the model size grows, indicating that a compact, core representation is learned from the early phase.  
(3) We provide initial evidence that ID-based rank selection in LoRA delivers efficiency gains without sacrificing perplexity, thereby highlighting the potential of ID-aware compression for large-scale NLP models.

%% file: section/sec2_related.tex
\label{sec:related}

In recent years, \textbf{Intrinsic Dimension (ID)} and \textbf{Local Intrinsic Dimensionality (LID)}~\cite{levina2004maximum,amsaleg2015estimating} have gained attention as indicators of the essential dimensionality of high-dimensional data. Since they capture the nonlinear manifold structure---beyond what linear methods like PCA can reveal---they provide valuable geometric insights into deeper feature representations. \citet{ansuini2019intrinsic} observed in the activations of CNNs that: (1) ID is smaller than the Euclidean dimension of each layer, (2) deeper layers tend to have a lower ID, and (3) higher ID often correlates with poorer generalization. For word embeddings, TwoNN~\cite{facco2017estimating} has shown that ID can compress to around 10 dimensions~\cite{ueda2024measuring}.

Meanwhile, low-rank approximation techniques such as LoRA~\cite{hulora} leverage the low-rank hypothesis to reduce inference and training costs for LLMs. LoRA freezes weights $W$ and learns a low-rank update $\Delta W = AB$ (with $A \in \mathbb{R}^{d \times r}$ and $B \in \mathbb{R}^{r \times k}$), where $r$ governs the compression--expressivity trade-off while drastically reducing trainable parameters. However, how far these representations can be compressed remains unexplored. Understanding this in embedding spaces is essential not only for deepening our grasp of representation learning but also for identifying new directions for model acceleration and memory efficiency.

%% file: section/sec3_method.tex
\label{sec:method}

We begin by describing how we estimate LID and ID, followed by three experiments that apply these methods to token embeddings.

\subsection{Method: LID and ID Estimation}
\label{subsec:LID_ID_estimation}

\textbf{Intrinsic Dimension Estimation.} 
Following ~\citet{levina2004maximum}, we estimate the \textbf{L}ocal \textbf{I}ntrinsic \textbf{D}imension (LID) of a point $x$ (e.g.\ one token embedding vector) via:
\vspace{-0.5\baselineskip}
\begin{equation}
\widehat{\mathrm{LID}}_k(x) 
= \left[
   \frac{1}{k-1} \sum_{i=1}^{k-1} \ln \frac{d_k(x)}{d_i(x)}
  \right]^{-1},
\label{eq:LID}
\end{equation}
where $d_i(x)$ is the distance from point $x$ to the $i$-th of total $k$ nearest neighbor ($k$ is a experiment hyper-parameter). Then, global ID~\cite{mackay2005comments} is computed as the harmonic mean of the LID across all $n$ embedding vectors:
\vspace{-0.5\baselineskip}
\begin{equation}
    \widehat{\mathrm{ID}}
    = \left[
       \frac{1}{n} \sum_{i=1}^{n} \widehat{\mathrm{LID}}_i^{-1}
      \right]^{-1}.
\label{eq:ID}
\end{equation}

\subsection{Experiment 1: ID Estimation for Word Embeddings}
\label{subsec:exp1}

Our first experiment evaluates whether widely used pre-trained word embeddings (Word2Vec~\cite{mikolov2013efficient}, GloVe~\cite{pennington2014glove}, FastText~\cite{bojanowski2017enriching}) occupy lower-dimensional manifolds than their ED. The models we use are available via the
Gensim library~\cite{rehurek_lrec}:
\lstinline|word2vec-google-news-300|,
\lstinline|glove-wiki-gigaword-300|,
and \lstinline|fasttext-wiki-news-subwords-300|. To assess the effect of linguistic structure, we also compare these embeddings to a random baseline consisting of vectors sampled from a normal distribution.

\vspace{0.1em}
\noindent \textbf{Experimental Procedure.} For each embedding vector from the full vocabulary, we use Euclidean distances and FAISS~\cite{douze2024faiss} to identify the $k=5$ nearest neighbors for the embedding vector, then compute $\widehat{\mathrm{LID}}_k$, as given in Eq.~\eqref{eq:LID}, and finally average these LID values to estimate the global ID following Eq.~\eqref{eq:ID}.

\vspace{0.2em}\noindent
\textbf{Random Baseline.}\quad
We generate 1e5 points from a $d$-dimensional Gaussian with mean $\mathbf{0}$ and covariance $\mathbf{I}$, where $d$ is set to equal with the extrinsic dimensionality of each evaluated embedding.

\begin{figure}[t]
    \centering
    \includegraphics[width=0.95\linewidth]{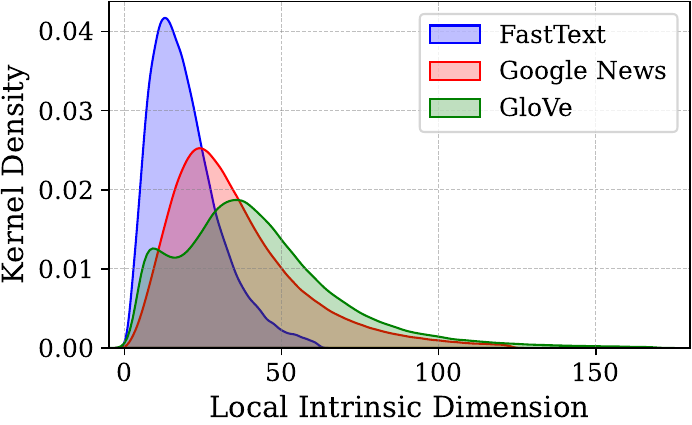}
    \vspace{-0.5\baselineskip}
    \caption{Kernel densities of LID values.}
    \label{fig:lid_hist}
    \vspace{-0.5\baselineskip}
\end{figure}

\vspace{0.3em}
\noindent
\textbf{Results.}\quad
Table~\ref{tab:lid_stats} shows the ID estimates obtained in this experiment, and Figure~\ref{fig:lid_hist} illustrates the distribution of LID values. The word embeddings yield IDs of about 10-30, significantly less than the \textit{random} baseline. Given that the ED is 300 for these vectors, the observed ID corresponds to approximately 3-10\% of the ED, suggesting strong redundancy in the original embedding dimension.

\begin{table}[t]
\centering
\caption{Estimated ID.}
\vspace{-0.5\baselineskip}
\label{tab:lid_stats}
\small
\resizebox{0.85\linewidth}{!}{
\begin{tabular}{lcccc}
\toprule
\textbf{Statistic} & \textbf{GloVe} & \textbf{FastText} & \textbf{Word2Vec} & \textbf{Random}\\
\midrule
ID & 24.77 & 13.19 & 24.75 & 130.3 \\
\bottomrule
\end{tabular}}
\end{table}

\subsection{Experiment 2: Redundancy Ratio Across Different LLM Scales}
\label{subsec:exp2}

Next, we measure the redundancy ratio in the embedding layer of the Pythia series~\cite{biderman2023pythia} with various scales from 14M to 12B parameters pre-trained under the same training data and conditions, to compare how the ID evolves at different scales under a consistent setting. For each model, let the extrinsic dimension be $\mathrm{ED}$, and let $\mathrm{ID}$ be the ID estimated by the method in \S\ref{subsec:LID_ID_estimation}, we define the \textit{redundancy ratio} as:
\begin{equation}
    \mathrm{Redundancy} 
    = \frac{\mathrm{ED} - \mathrm{ID}}{\mathrm{ED}},
\label{eq:redundancy}
\end{equation}
and observe it among various model scales. Unlike \S\ref{subsec:exp1}, we focus on this ratio instead of $\mathrm{ED}$, since $\mathrm{ED}$ varies across models.

\begin{figure}[t]
    \centering
    \includegraphics[width=0.9\linewidth]{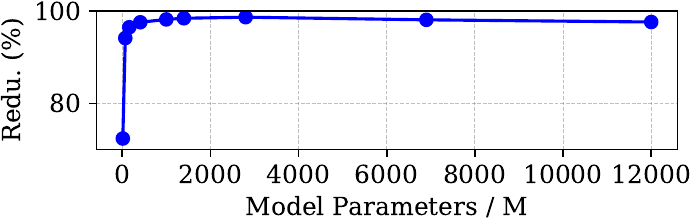}
    \vspace{-0.5\baselineskip}
    \caption{Redundancy Ratio against model parameters.}
    \label{fig:red_aginst_para}
    \vspace{-0.5\baselineskip}
\end{figure}

\vspace{0.2em} \noindent
\textbf{Results.}\quad
Table.~\ref{tab:pythia_redundancy} and Figure.~\ref{fig:red_aginst_para} present the results of redundancy ratios. As the model size grows, the ID also increases, yet the redundancy ratio remains very high, between roughly 90\% and 98\%. Moreover, from \verb|pythia-410m| onward, the redundancy ratio stabilizes at around 98\%. In other words, for sufficiently large models, the redundancy ratio does not undergo significant change.

\begin{table}[t]
\centering
\caption{Redundancy Ratio (Redu.\ (\%)) alongside ID and ED for Pythia models with various scales.}
\vspace{-0.5\baselineskip}
\small
\label{tab:pythia_redundancy}
\begin{tabular}{lccc}
\toprule
\textbf{Model} & \textbf{Redu.\ (\%)} & \textbf{ID}& \textbf{ED} \\ \midrule
pythia-14m & 72.40 & 35.33 & 128 \\ 
pythia-70m & 94.14 & 29.99 & 512 \\ 
pythia-160m & 96.49 & 26.97 & 768 \\ 
pythia-410m & 97.56 & 24.95 & 1024 \\ 
pythia-1b & 98.18 & 37.23 & 2048 \\ 
pythia-1.4b & 98.43 & 32.20 & 2048 \\ 
pythia-2.8b & 98.66 & 34.18 & 2560 \\ 
pythia-6.9b & 98.09 & 78.30 & 4096 \\
pythia-12b & 97.62 & 121.82 & 5120 \\ \bottomrule
\end{tabular}
\end{table}

\subsection{Experiment 3: ID Estimation During LLM Training}
\label{subsec:exp3}

To examine how the embedding space of LLMs evolves during training, we utilize the model checkpoints periodically saved along the training dynamics from 1e3 to 1e4 steps at intervals of 1e3, and from 1e4 to 1.43e5 steps at intervals of 5e3. At each checkpoint, we estimate $\widehat{\mathrm{LID}}_k$ using Eq.~\eqref{eq:LID}, $\widehat{\mathrm{ID}}$ using Eq.~\eqref{eq:ID} thereby tracking changes in ID throughout training. Due to limited GPU resources, we restrict our experiments to models ranging from \verb|pythia-14m| to \verb|pythia-1.4b|.

\begin{figure}[t] 
    \centering
    \includegraphics[width=0.95\linewidth]{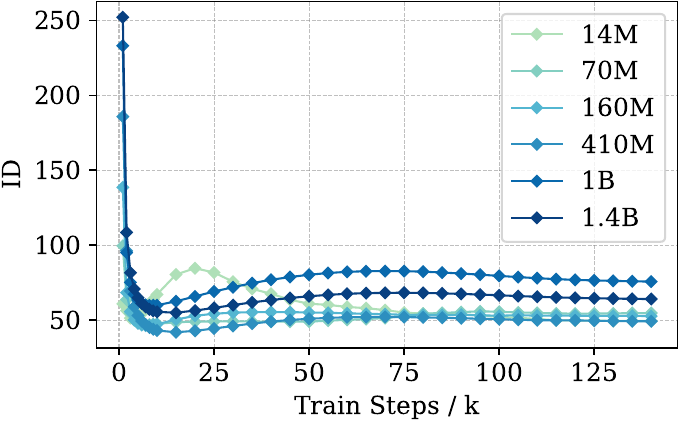}
    \vspace{-0.5\baselineskip}
    \caption{Dynamics of ID against the training steps.}
    \label{fig:process}
    \vspace{-0.5\baselineskip}
\end{figure}

\noindent
\textbf{Results.}\quad
\noindent
Figure.~\ref{fig:process} presents our findings. We observe a sharp decline in ID during the initial training stages, followed by a more gradual convergence. The smallest model, \verb|pythia-14m|, exhibits relatively unstable behavior, which is generally acceptable for smaller-scale models~\cite{tirumala2022memorizationoverfittinganalyzingtraining}.

\subsection{Experiment 4: LoRA with ID-driven Rank Choice}
\label{subsec: LoRA}

\begin{figure}[t] 
    \centering
    \includegraphics[width=0.43\textwidth]{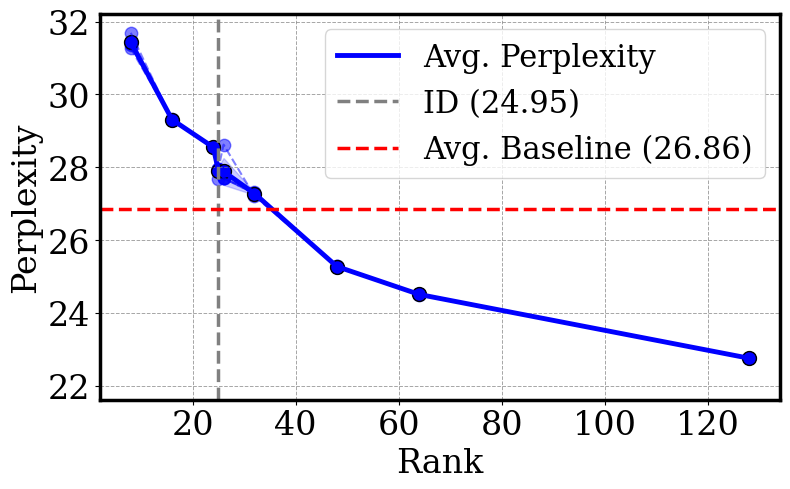}
    \caption{Validation perplexity against LoRA inner dimensions on \texttt{pythia-410m}.}
    \label{fig:LoRA}
    \vspace{-1\baselineskip}
\end{figure}

In \S\ref{subsec:exp2}, we obtained the ID of each model’s embedding layer and used it to guide the rank (inner dimension) selection of LoRA~\cite{hulora}. Similarly to before, we apply LoRA \emph{only} to the embedding layer (e.g., \verb|gpt_neox.embed_in| in Pythia) for a causal language modeling task on the WikiText-2 dataset, where the dataset is tokenized to a maximum sequence length of 256 and any empty samples are discarded. We systematically vary the LoRA rank 
\{8, 16, 24, 25, 26, 32, 48, 64, 128\} around the estimated ID ($\sim$ 24.95), and train only the LoRA parameters on the aforementioned object for 5 epochs with a per-device batch size of 32, and compute the perplexity on the validation set as $\mathrm{exp}(\mathrm{loss})$ to evaluate the effect of LoRA. This setup allows us to examine how closely the optimal LoRA rank aligns with the ID, as well as whether ranks below or above the ID threshold significantly affect the model’s performance.


\noindent
\textbf{Results.}\quad
\noindent
Figure.~\ref{fig:LoRA} presents our findings. In Figure.~\ref{fig:LoRA}, error bands corresponding to $\pm \sigma$ are displayed. We find that in LoRA, ranks below the ID lead to a clear performance drop, whereas ranks above the ID improve results slightly. Around the ID, performance jumps sharply before declining again, suggesting that ID is pivotal for balancing compactness and capacity in LoRA.

%% file: section/sec4_discussion.tex
\label{sec:discussion}
\subsection{RQ1: The Gap between ED and ID is Significant}
\label{subsec:rq1}

Word embeddings with an ED of 300 typically exhibit an ID of around 10-30, which aligns with the findings on the sentence embedding (\textit{activation}) of~\citet{ueda2024measuring}. 
It can be inferred that language prior leads the embeddings and also activations to appear more structured and low-ID geometries, compared to random vectors.

Notably, FastText embeddings exhibit a significantly lower ID compared to those from other models. This phenomenon may be attributed to FastText’s subword segmentation, with additional contributions potentially coming from factors such as the training data and token frequency. To investigate this, we conducted a preliminary experiment with various tokenizers to assess how different tokenization strategies affect the resulting ID. Details and results are provided in the Appendix~\ref{sec:appendix}.

\subsection{RQ2: Redundancy Ratio Persists at a High Level}
\label{subsec:rq2}

In Fig.~\ref{tab:pythia_redundancy}, our scale-based analysis reveals that as the model size grows, the ID also increases but still lags significantly behind the ED, resulting in about a 98\% redundancy ratio. This suggests that many dimensions remain underutilized, even though large models offer ample representational capacity. Moreover, high redundancy may, in fact, mirror the inherent complexity of language, providing nuanced flexibility for downstream tasks and cautioning against viewing it as purely inefficiency. In detail, it can be considered that during the fine-tuning onto a downstream task, the model can enable the unused dimensionalities as a ``channel'' for the related information.

Additionally, \S\ref{subsec:exp3} shows that early training rapidly finds a compact, low-dimensional representation of core linguistic features, followed by a slower phase of refinement. 

\noindent
\textbf{Possible Explanations for the Rapid Emergence of Low-Dimensional Structure.} We conjecture that the embedding layer quickly converges to a low-dimensional manifold due to the over-parameterized nature of the model and the intrinsic clustering in natural language. Specifically, during the initial training phase, frequent tokens are rapidly grouped in a semantically meaningful subspace, while infrequent tokens remain scattered around the periphery, effectively reducing the global degrees of freedom. This phenomenon aligns with previous work on Neural Collapse~\cite{gaorepresentation, cho-etal-2025-understanding} in classification settings, suggesting that early training emphasizes global structure. Moreover, the manifold hypothesis posits that real-world data often lie on a low-dimensional manifold; our ID estimation lends empirical support to this claim in the context of large-scale language models. In later stages of training, ID remains relatively stable, indicating a phase where the primary geometry is refined rather than fundamentally restructured. We believe that additional factors such as learning rate schedules, token frequency distributions (Zipf’s law), and subword segmentation might further influence the speed and extent of ID convergence. Future work will include in-depth analyses of these factors and their interplay with optimization dynamics.

%% file: section/sec5_conclusion.tex
\noindent
We have shown that while embeddings in both small and large models nominally span hundreds or thousands of dimensions, their \textbf{effective} dimensionality, ID, is remarkably low. Notably, ID emerges early in training and remains far below the ED, leaving significant redundancy. Crucially, these findings inform practical compression strategies such as LoRA, where selecting a rank close to the ID can preserve performance while reducing parameters. In short, the ID-based perspective offers both theoretical insight into LLM embeddings and a concrete path toward more efficient, scalable model deployment.

\paragraph{Future Work.}
We plan to explore ID in additional layers and architectures, extend our approach to cross-linguistic and diachronic corpora, and further investigate ID-based compression methods to enhance LLM interpretability and performance.

%% file: section/limitation.tex
One limitation of this study is that it focuses exclusively on the Pythia model, thereby restricting the generalizability of our findings to other architectures. Additionally, due to the practical constraints posed by our available GPU resources, the experimental scale remains somewhat smaller compared to contemporary large-scale language models. Consequently, caution should be exercised when extrapolating these results to larger or more diverse model families.

%% file: section/Appendix.tex
To examine how subword segmentation, training data, or frequency characteristics might influence the ID, we trained word embeddings using various tokenizers on the AGNews corpus~\cite{zhang2015character}. Specifically, we compared SentencePiece, \textbf{B}yte-\textbf{P}air \textbf{E}ncoding (BPE), and whitespace tokenization (WS) under both Word2Vec and FastText frameworks. Table~\ref{tab:tokenizer} lists the resulting ID values for embeddings with an ED of 300, using a vocabulary sample of 10,000 tokens.

\begin{table}[t]
\centering
\caption{ID for Each Tokenizer (ED = 300, Vocab Sample = 10,000). WS indicates whitespace tokenization.}
\label{tab:tokenizer}
\begin{tabular}{lc}
\toprule
\textbf{Tokenizer} & \textbf{ID} \\ 
\midrule
Word2Vec-SentencePiece & 24.7846 \\ 
Word2Vec-BPE           & 24.7275 \\ 
Word2Vec-WS            & 27.0036 \\ 
FastText-SentencePiece & 11.3744 \\ 
FastText-BPE           & 11.9805 \\ 
FastText-WS            & 10.7492 \\ 
\bottomrule
\end{tabular}
\end{table}

We observe that FastText embeddings generally yield lower ID values than Word2Vec across all tokenizers, suggesting that subword-level modeling may help reduce the intrinsic dimensionality. However, further analysis is needed to confirm whether these differences are indeed due to segmentation approaches, data frequency characteristics, or training hyperparameters.